\documentclass{article}

\usepackage[preprint]{corl_2026} 
\usepackage{wrapfig}
\usepackage[nolist]{acronym} 
\usepackage{amsfonts,siunitx,booktabs,cite} 
\usepackage{graphics} 
\usepackage{epsfig} 
\usepackage{times} 
\usepackage{amsmath} 
\usepackage{amssymb}  
\usepackage[linesnumbered,ruled]{algorithm2e}
\SetKw{Continue}{continue}
\usepackage{stmaryrd}
\usepackage{soul}
\usepackage{bm}

\usepackage{amsthm}
\usepackage[inline]{enumitem}
\usepackage{graphicx}
\usepackage{caption}

\usepackage{subcaption}
\usepackage{multirow}
\usepackage{enumitem}
\usepackage{svg}
\usepackage{float}
\usepackage[normalem]{ulem}
\usepackage[table,x11names,dvipsnames]{xcolor}
\definecolor{salmon}{RGB}{250, 128, 114}
\usepackage{hyperref}
\usepackage{algpseudocode}

\usepackage{scalerel}
\usepackage{tikz,standalone,pgfplots}
\pgfplotsset{compat=1.18}
\usepackage{pifont}
\usepackage{overpic}
\usepackage{balance}
\usepackage[normalem]{ulem} 

\title{MoDex: A Diffusion Policy for Sequential Multi-Object Dexterous Grasping}

\author{
     Haofei Lu $^{\dag 1}$,
     Hongjia Liu $^{1}$,
     Yifei Dong $^{1}$,
     Florian T. Pokorny $^{1}$,
     Jens Lundell $^{2}$,
     Danica Kragic $^{1}$\\[1.5ex]
     $^{1}$Department of Robotics, Perception and Learning, KTH Royal Institute of Technology, Sweden.\\
     $^{2}$Robotics and Autonomous Systems at University of Turku, Finland.\\
    \texttt{\{haofeil,hongjial,yifeid,fpokorny,dani\}@kth.se, jens.lundell@utu.fi}
}

\begin{document}
\maketitle
\newacro{dnn}[DNN]{deep neural network}
\newacro{fcn}[FCN]{fully convolutional network}
\newacro{sdf}[SDF]{signed distance function}
\newacro{cnn}[CNN]{convolutional neural network}
\newacro{gnn}[GNN]{graph neural network}
\newacro{dl}[DL]{deep learning}
\newacro{ml}[ML]{machine learning}
\newacro{mc}[MC]{Monte Carlo}
\newacro{mlp}[MLP]{multi-layer perceptron}
\newacro{dof}[DoF]{degree of freedom}
\newacroplural{dof}[DoFs]{degrees of freedom}
\newacro{vlm}[VLM]{vision language model}
\newacro{vae}[VAE]{variational autoencoder}
\newacro{cvae}[CVAE]{conditional variational autoencoder}
\newacro{methodname}[VCGS]{Variational Constrained Grasp Sampler}
\newacro{fps}[FPS]{farthest point sampling}
\newacro{tai}[TaI]{target as input}
\newacro{pca}[PCA]{principal component analysis}
\newacro{pc}[PC]{principal component}
\newacro{auc}[AUC]{area under the curve}
\newacro{elbo}[ELBO]{evidence lower bound}
\newacro{bps}[BPS]{basis point set}
\newacro{mala}[MALA]{Metropolis-adjusted Langevin algorithm}
\newacro{dp}[DP]{diffusion policy}
\newacro{il}[IL]{imitation learning}
\newacro{rl}[RL]{reinforcement learning}
\newacro{bc}[BC]{behaviour cloning}
\newacro{os}[OS]{opposition space}
\newacroplural{os}[OSes]{opposition spaces}
\newacro{mcp}[MCP]{metacarpophalangeal}
\newacro{gan}[GAN]{generative adversarial network}
\newacro{dp3}[DP3]{3D Diffusion Policy}
\newacro{dppo}[DPPO]{diffusion policy policy optimization}
\newacro{ppo}[PPO]{proximal policy optimization}
\newacro{ddim}[DDIM]{denoising diffusion implicit model}
\newacro{ddpm}[DDPM]{denoising diffusion probabilistic model}
\newacro{mdp}[MDP]{Markov decision process}
\newacro{osc}[OSC]{operational space controller}
\newacro{rrt}[RRT]{rapidly exploring random tree}
\newacro{vla}[VLA]{vision-language-action}
\newcommand{\jensrmk}[1]{{\color{blue} {[\bf j: #1]}}}
\newcommand{\jens}[1]{{\color{blue}#1}}
\newcommand{\zehangrmk}[1]{{\color{green} {[\bf f: #1]}}}
\newcommand{\zehang}[1]{{\color{green}#1}}
\newcommand{\haofeirmk}[1]{{\color{cyan} {[\bf v: #1]}}}
\newcommand{\haofei}[1]{{\color{cyan}#1}}
\newcommand{\yifeirmk}[1]{{\color{purple} {[\bf v: #1]}}}
\newcommand{\yifei}[1]{{\color{purple}#1}}

\definecolor{drakgreen}{rgb}{0.0, 0.5, 0.0}  

\newcommand{\equationref}[1]{\hyperref[#1]{Eq.~(\ref*{#1})}}
\newcommand{\figref}[1]{\hyperref[#1]{Fig.~\ref*{#1}}}
\newcommand{\tabref}[1]{\hyperref[#1]{Table~\ref*{#1}}}
\newcommand{\secref}[1]{\hyperref[#1]{Section~\ref*{#1}}}
\newcommand{\algoref}[1]{\hyperref[#1]{Algorithm~\ref*{#1}}}
\newcommand{\linesref}[2]{Lines~\ref{#1}-\ref{#2}}
\newcommand{\defref}[1]{\hyperref[#1]{Definition~\ref*{#1}}}
\newcommand{\figsref}[2]{Figures~\ref{#1}-\ref{#2}}
\newcommand{\subfigref}[1]{(\subref{#1})}

\newcommand{\norm}[1]{\left\lVert#1\right\rVert}
\newcommand{\Var}{\mathrm{Var}}
\newcommand{\ra}[1]{\renewcommand{\arraystretch}{#1}}
\newcommand{\tbs}[1]{\renewcommand{\tabcolsep}{#1pt}}
\newcommand{\abs}[1]{\left\lvert#1\right\rvert}

\newcommand{\matr}[1]{\mathbf{#1}}
\newcommand{\argmax}{\operatornamewithlimits{argmax}}
\newcommand{\argmin}{\operatornamewithlimits{argmin}}
\newcommand*{\prob}{\mathsf{P}}
\newcommand{\de}[1]{\operatorname{d}\!#1}
\newcommand{\etal}[1]{#1 et al.}
\newcommand{\set}[1]{\mathcal{#1}}
\newcommand{\clip}{\operatorname{clip}}
\newcommand{\OS}{\mathrm{OS}}

\def\methodname{MoDex}
\def\diffusionname{SeqDiffuser}
\def\methodshort{SeqG}
\def\multigraspshort{MulG}
\def\diffusionshort{SeqD}

\def\datasetname{SeqDataset}
\def\graspemshort{G'Em}
\def\graspem{Grasp'Em}
\def\datasetshort{SData}
\def\franka{Franka Emika Panda}
\def\kinect{Kinect 3.0}
\def\multigrasp{MultiGrasp}
\def\graspnetta{GraspNet \ac{tai}}
\def\doflong{degrees of freedom}
\def\dof{DoFs}

\def\bestcolor{(best viewed in color)}
\def\sota{state-of-the-art}
\def\ie{, \textit{i.e.}, }
\def\eg{\textit{e.g.}, }
\def\pc{point cloud}
\def\pcs{point clouds}
\def\epst{\multicolumn{1}{c}{$\epsilon$}}
\def\vt{\multicolumn{1}{c}{$v$}}
\def\nat{\multicolumn{1}{c}{--}}
\newcommand{\blue}[1]{\textcolor{black}{#1}}
\def\pointnet{PointNet++~\cite{qiPointNetDeepHierarchical2017a}}

\begin{figure}[htbp]
    \centering
    \includegraphics[width=\textwidth]{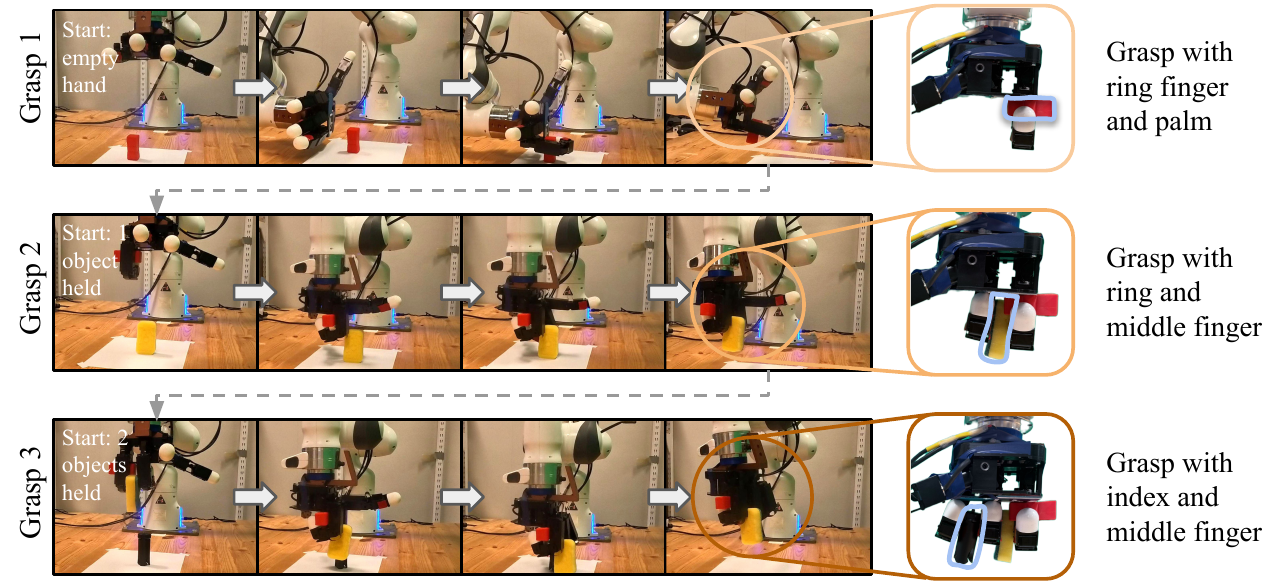} 
    \caption{\methodname{} \emph{sequentially} picks three objects with a single dexterous hand, securely holding all previously grasped objects while picking the next one. All grasps are produced by a single policy. Dashed arrows link the end of one grasp to the start of the next; close-ups (right) show the final hand configuration after each. }
    \label{fig:teaser}
\end{figure}

\begin{abstract}
This work addresses sequentially grasping multiple objects with a single dexterous hand without releasing those already held. Most dexterous grasping methods commit all of the hand's \aclp{dof} to a single object, underutilizing its dexterity and leaving no redundancy for subsequent grasps. The proposed solution, \methodname{}, is a \acl{dp} that predicts the next gripper pose directly from observations, conditioned on an opposition space and point cloud. The \acl{os} condition specifies which fingers participate in the current grasp, enabling the gripper to use only a subset of its available \aclp{dof} while reserving the remaining \aclp{dof} for subsequent grasps. To facilitate sim-to-real transfer, \methodname{} is trained in two stages: first through \acl{il} on expert demonstrations, and subsequently through reinforcement learning fine-tuning, which consistently improves success rates over the pre-trained policy. We evaluate \methodname{} in simulation on a MuJoCo-based \franka{} robot equipped with an Allegro Hand and on the corresponding real-world hardware platform. Across both simulation and real-world experiments, \methodname{} achieves higher success rates than the evaluated learning-based baselines, improving performance by 2.92–17.92\% and 6.67–17.78\%, respectively. Project page: \url{https://modex2026.github.io/}.



\end{abstract}

\keywords{Dexterous Grasping, Robot Manipulation, Reinforcement Learning} 

\section{Introduction}
\label{sec:Introduction}
To date, most dexterous grasping methods use all the end-effector's \acp{dof} for picking objects~\citep{graspxl, zhang2025RobustDexGrasp, dexdiffuser}. However, these methods rarely achieve higher grasp success rates than approaches using low \ac{dof} parallel-jaw grippers~\citep{anygrasp}. Consequently, some argue that a dexterous arm paired with a simple gripper is sufficient for most manipulation tasks~\citep{6298887}. But what if the problem is that current methods underutilize the gripper's available dexterity? Motivated to answer this question, we propose \methodname{}, the first \ac{dp}~\citep{chi2024diffusionpolicy} for sequentially picking multiple objects with a dexterous gripper, as demonstrated in \figref{fig:teaser}.

\methodname{} is trained to map partial point clouds of the scene, previously executed grasps, and an \ac{os} descriptor that specifies which fingers participate in the current grasp~\citep{feix2015grasp} to the next desired end-effector pose. This design lets \methodname{} sequentially grasp multiple objects directly from raw perception, unlike previous multi-object grasping methods that either require known object models~\citep{seqgrasp}, decoupled grasp generation and execution~\citep{yao2023exploiting}, or near-fixed object poses~\citep{he2025sequential}. \methodname{} is pre-trained on a new dataset containing  thousands of expert sequential grasping demonstrations generated automatically by a heuristic pose generator and a controller in simulation. To extend the rather narrow state-action distribution covered in pre-training, \methodname{} is fine-tuned using the  \ac{rl} framework \ac{dppo}~\citep{ren2025diffusion} on a reward specifically designed for sequential multi-object grasping. \methodname{} is benchmarked against representative \ac{il} and \ac{rl} baselines on 15 objects of diverse geometry on a simulated \franka{} arm equipped with an Allegro Hand. We also conduct 90 real-world grasping sequences using the same physical hardware setup. The results demonstrate that \methodname{} outperforms \ac{dp3} by 2.92–17.92\% in simulation and 6.67-17.78\% in the real world.

In summary, our contributions are:
\begin{itemize}
    \item \methodname{}, the first \ac{dp} for sequential multi-object grasping (\secref{sec:method}).
    \item An algorithm for collecting expert sequential multi-object grasping trajectories (\secref{sec:dataset}). 
    \item An extensive experimental evaluation exploring the strengths and limitations of \methodname{} compared to \ac{il} and \ac{rl} baselines (\secref{sec:experiments}).
\end{itemize}


\section{Related Work}

Our work spans dexterous and multi-object grasping, as well as policy learning. Below, we review each of these separately.
\label{sec:relatedwork}

\paragraph{Dexterous Grasping.} Traditionally, dexterous robotic grasps were generated by optimizing them with respect to a surrogate grasp quality metric, such as force closure~\citep{dexgraspnet, dfc,graspit,eigengrasp,yin2025lightninggrasphighperformance}. While these methods provide theoretical guarantees, they assume known object models and often struggle with the ``sim-to-real'' gap due to unmodeled environmental uncertainties. To address these limitations, researchers turned to data-driven models trained to directly map partial observations to high-quality static grasp poses~\citep{dexdiffuser,lu2023ugg,jensdgcc, wei2024dro,mayer2022ffhnet,zhang2024dexgraspnet}. Other, more recent methods~\citep{graspxl,zhang2025RobustDexGrasp, makarova2026diffusionrlefficienttrainingdiffusion} train \ac{il} or \ac{rl} policies to output continuous grasping actions rather than static grasp poses. Still, single-object grasping methods fail to fully exploit the high \ac{dof} potential of dexterous hands.

\paragraph{Multi-Object Grasping.} Research into dexterous multi-object grasping is still in its early stages. Initial efforts~\citep{multigrasp,sun2022multi} have focused on \emph{simultaneous} multi-object grasping to improve efficiency and reduce execution time by avoiding repeated arm repositioning. However, these methods require objects to be in close proximity and do not exploit the full dexterity of the gripper, as multi-object grasping is best solved by first enveloping all the objects with the hand and then closing all fingers simultaneously. More recent works~\citep{yao2023exploiting,seqgrasp,he2025sequential} have explored the \emph{sequential} multi-object grasping problem. To solve this problem, the number of fingers used to pick an object at a time should be minimized, ensuring that as many \acp{dof} as possible are available for subsequent grasps. However, real-world deployment remains challenging for existing sequential multi-object grasping methods, which commonly rely on collision-free path planning~\citep{yao2023exploiting, seqgrasp}. This work overcomes these limitations by training a single multi-object grasping policy that directly maps incomplete object observations to grasp actions.

\paragraph{Diffusion Policies.} Diffusion models were recently applied to learning visuomotor policies~\citep{chi2024diffusionpolicy}. These methods, called \acp{dp}, treat action generation as an iterative denoising process. The main benefits over other policy learning frameworks are that \acp{dp} capture multimodal action distributions more faithfully and are more stable to train~\citep{chi2024diffusionpolicy,ren2025diffusion}. Therefore, many \ac{dp} extensions have been proposed, such as incorporating 3D point cloud observations to improve spatial reasoning~\citep{Ze2024DP3}, adding \ac{rl} fine-tuning to overcome the narrow state-action coverage of behavior cloning~\citep{ren2025diffusion}, and using it as the action head in modern \acl{vla} models~\citep{trilbmteam2025carefulexaminationlargebehavior, gr00tn1_2025,li2024cogact,liu2025rdt,wendexvla}. However, to the best of our knowledge, \acp{dp} have not been used to learn sequential multi-object grasping policies, which is the problem addressed in this work.


\section{Problem Formulation}
\label{sec:problemformulation}

In sequential multi-object grasping, the problem is to grasp the $i$-th target object while holding all the previously $i{-}1$ objects grasped, where $i \in \{1, \ldots, I\}$ is referred to as a grasp stage and $I$ is the maximum number of objects that the gripper can possibly grasp. The kinematic structure of the Allegro hand, which is used in this work, restricts $I \leq 3$. The goal then is to learn a control policy ${\pi_{\theta}:\mathcal{O}\rightarrow\mathcal{A}^{T_p}}$ that maps the current observation $\mathbf{o}_{t, i}\in\mathcal{O}$ to an action chunk ${\mathbf{A}_t=\left[\mathbf{a}_t,\mathbf{a}_{t+1},\ldots,\mathbf{a}_{t+T_p-1}\right]\in\mathcal{A}^{T_p}}$ of prediction horizon $T_p$, with the first $T_a \leq T_p$ actions executed on the robot before the policy is queried again. Throughout, $t$ denotes the environment timestep within grasp stage $i$.

Observations are represented as the tuple ${\mathbf{o}_{t,i}=\{\mathbf{f}_i,~\mathbf{h}_{i-1},~\mathbf{q}_t,~\mathcal{P}_t\}}$, where $\mathbf{f}_i$ is the \ac{os} vector for the $i$-th grasp, $\mathbf{h}_{i-1}$ is a grasp history,~$\mathbf{q}_t\in \mathbb{R}^{D+J}$ is the robot proprioception stacking the $D$ arm joint angles and the $J$ hand joint angles, and $\mathcal{P}_t = \{\mathbf{x}_j\}_{j=1}^{M}$, with $\mathbf{x}_j \in \mathbb{R}^3$, is a partial object point cloud of the environment. The \ac{os} vector $\mathbf{f}_i$ describes which \acp{dof} should be used to pick the current target object~\citep{seqgrasp}, while the grasp history $\mathbf{h}_{i-1}$ describes which \acp{dof} are already used for picking the previous $i-1$ objects. Each action in the action chunk is represented as the vector ${\mathbf{a}_t=[\mathbf{p}_t, \mathbf{r}_t, \boldsymbol{\phi}_t]\in\mathbb{R}^{6+J}}$, where $\mathbf{p}_t\in\mathbb{R}^3$ is the hand base position, $\mathbf{r}_t\in\mathbb{R}^3$ is the hand base orientation in axis-angle form, and $\boldsymbol{\phi}_t\in\mathbb{R}^J$
denotes the hand joint angles. In this work, $D=7$ and $J=16$ for the Franka Panda arm and Allegro Hand, respectively. The problem then becomes: 1) how to mathematically represent $\mathbf{f}_i$ and $\mathbf{h}_{i-1}$, and 2) how to train the policy $\pi_{\theta}$ from data.



\section{Preliminaries}
\label{sec:preliminaries}

This section reviews the background on \acp{dp}~\citep{chi2024diffusionpolicy} and \ac{dppo}~\citep{ren2025diffusion} necessary for training the sequential multi-object grasping policy $\pi_\theta$. 

\subsection{Diffusion Policies}
A \ac{dp}~\citep{chi2024diffusionpolicy} parameterizes the policy
$\pi_\theta$ as a \ac{ddpm}~\citep{ho2020denoising} that denoises action chunks of horizon $T_p$. Starting from Gaussian noise $\mathbf{A}_t^K \sim \mathcal{N}(\mathbf{0}, \mathbf{I})$, a \ac{dp} iteratively denoises the chunk over $K$ steps,
\begin{equation}
    \mathbf{A}_t^{k-1} \sim p_\theta(\mathbf{A}_t^{k-1} \mid \mathbf{A}_t^k, \mathbf{o}_{t,i}) := \mathcal{N}\!\left(\mathbf{A}_t^{k-1};\,
    \mu_k(\mathbf{A}_t^k, \varepsilon_\theta(\mathbf{A}_t^k, \mathbf{o}_{t,i}, k)),\, \sigma_k^2 \mathbf{I}\right),
\end{equation}
where $\varepsilon_\theta$ is a neural network predicting the noise at step $k$ and
$\sigma_k^2$ follows a fixed schedule. After denoising, the first $T_a \leq T_p$ actions of the clean chunk $\mathbf{A}_t^{0}$ are executed on the robot before the policy is queried again. The network $\varepsilon_\theta$ is pre-trained via \ac{il} using the behavior cloning objective:
\begin{equation}
    \label{eq:behavior_cloning_objective}
    \mathcal{L}_{\mathrm{BC}}(\theta) = \mathbb{E}_{(\mathbf{o}_{t,i},\, \mathbf{A}^0_t) \sim
    \mathcal{D}_{\mathrm{off}},\; k,\; \varepsilon^k}\left[\|\varepsilon^k - \varepsilon_\theta(\mathbf{A}^0_t + \varepsilon^k,\, \mathbf{o}_{t,i},\, k)\|^2\right],
\end{equation}
where $\mathcal{D}_{\mathrm{off}}$ is a dataset of trajectories, $k$ is a uniformly sampled denoising step, and $\varepsilon^k \sim \mathcal{N}(\mathbf{0}, \mathbf{I})$
is the noise injected at that step.

\subsection{Diffusion Policy Policy Optimization}
A major limitation of \ac{dp} and behavior cloning policies in general is that their robustness to out-of-distribution states is limited, as the training data cover a rather narrow state-action distribution~\citep{osa2018algorithmic}. Thus, to make policies more robust, a common practice is to refine them via \ac{rl} fine-tuning~\citep {ding2024diffusionbased}. For \acp{dp}, the primary \ac{rl} finetuning method is \ac{dppo}~\citep{ren2025diffusion},
which treats the $K$-step denoising process as an \ac{mdp} nested within the environment \ac{mdp}. Based on this two layer \ac{mdp}, \ac{dppo} formulates the policy gradient updates with the \ac{ppo}~\citep{schulman2017proximal} objective:
\begin{equation}
    \label{eq:dppo}
    \mathcal{L}_\theta = \mathbb{E}_{\mathcal{D}_{\mathrm{itr}}} \left[ \min\!\left(
    \hat{A}_{\bar t} \frac{\bar{\pi}_\theta}{\bar{\pi}_{\theta_{\mathrm{old}}}},\;
    \hat{A}_{\bar t} \cdot \clip\!\left(\frac{\bar{\pi}_\theta}
    {\bar{\pi}_{\theta_{\mathrm{old}}}}, 1{-}\epsilon_{\mathrm{clip}}, 1{+}\epsilon_{\mathrm{clip}}\right)
    \right) \right],
\end{equation}
where $\hat{A}_{\bar{t}} = \gamma_{\text{denoise}}^{k}\bigl(\bar{r} - 
\tilde{V}(\mathbf{o}_{t,i})\bigr)$ is the denoising-discounted advantage. The discount ${\gamma_{\text{denoise}} \in (0, 1)}$ downweights the gradient signal at noisier, earlier denoising steps (large $k$), where the predicted action carries little task-relevant information, and concentrates learning on the later steps (small $k$) that most directly shape 
the executed action. The return $\bar{r}$ is the environment return associated with the rollout, and $\tilde{V}(\mathbf{o}_{t,i})$ is the value function. To improve efficiency, only the last $K' \leq K$ denoising steps are fine-tuned using \acl{ddim} sampling~\citep{songdenoising}.

\section{\methodname{}}
\label{sec:method}

\begin{figure}[tp]
    \centering
    \includegraphics[width=0.9\textwidth]{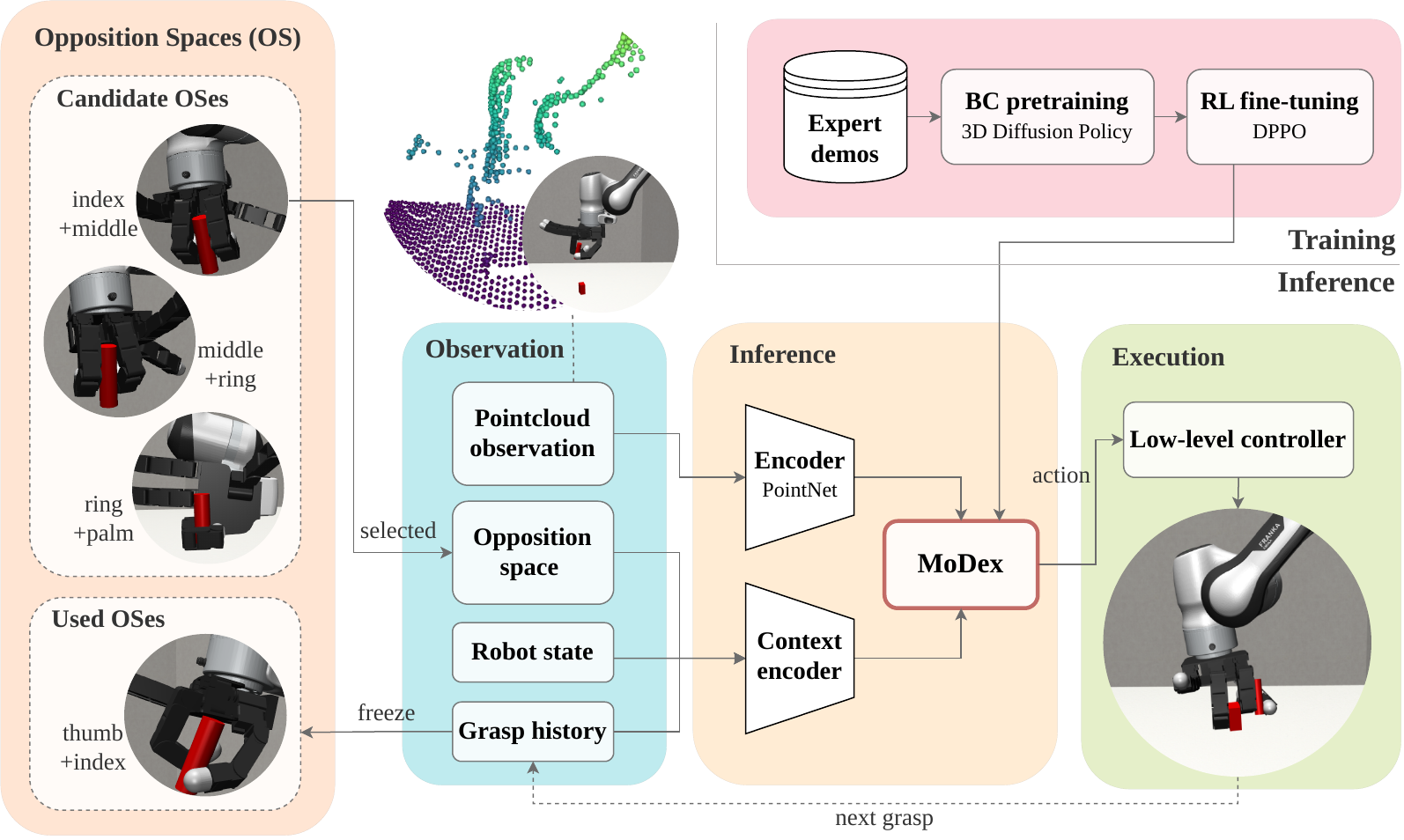}
    \caption{\textbf{Method overview.} \methodname{} maps an observation, including a point cloud, the selected \ac{os}, the robot state, and the grasp history of \acp{os} already used (bottom left), to the next grasp action. A PointNet encoder and a context encoder feed the diffusion policy. The executed grasp is appended to the history before the next object. Training (top right): the policy is first pre-trained by behavior cloning on expert demonstrations (\ac{dp3}), then RL fine-tuned with DPPO.}
    \label{fig:methodoverview}
\end{figure}

\figref{fig:methodoverview} shows an overview of \methodname{}. The sections below detail our specific contributions: how to architect the sequential multi-object grasping \ac{dp}, how \acp{os} and grasp history are mathematically represented and integrated into the policy architecture, and which rewards are needed for successful \ac{rl} fine-tuning.

\subsection{\texorpdfstring{\ac{os}}{OS} and Grasp History Context}
\label{subsec:grasp_history}

A central challenge in single-policy sequential multi-object grasping is producing qualitatively distinct grasps that use as few \acp{dof} as possible for each successive object while leaving the \ac{dof} used in prior grasps untouched. We address this through two conditioning signals: the \emph{\ac{os}} and the \emph{Grasp History Context}.

\paragraph{Opposition Space.}
Originally proposed by~\citet{feix2015grasp}, an \ac{os} characterizes a grasp
by the set of finger pairs acting in opposition during object contact.
We represent the \ac{os} for the $i$-th grasp as a binary indicator vector $\mathbf{f}_i \in \{0,1\}^L$, where $L$ is the number of fingers in the hand ($L=4$ for the Allegro Hand) and $(\mathbf{f}_i)_j = 1$ iff finger $j$ participates in the grasp. In this work, we restrict the \acp{os} that the Allegro Hand can achieve to the four shown in \figref{fig:methodoverview}. For example, $\mathbf{f}_1 = [1,1,0,0]$ engages only the thumb and index finger, leaving the remaining fingers free for subsequent grasps. We include $\mathbf{f}_i$ into \methodname{} by concatenating it, at every denoising step $k$, to each action in the noisy action chunk $\mathbf{A}_t^{k}$ , forming the \ac{os}-augmented action chunk ${\tilde{\mathbf{A}}_t^{k}=\left[\mathbf{a}^k_t\;\|\;\mathbf{f}_i,~\mathbf{a}^k_{t+1}\;\|\;\mathbf{f}_i,~\ldots,~\mathbf{a}^k_{t+T_p-1}\;\|\;\mathbf{f}_i\right]}$.

\paragraph{Grasp History Context.}
The condition $\mathbf{f}_i$ only tells which \acp{dof} should be used to grasp the current object, but ignores which fingers have already been used by prior grasps. We therefore
introduce the Grasp History Context, defined as the vector ${\mathbf{h}_{i-1} = \left[\mathbf{f}_1 \;\|\; \cdots \;\|\;
    \mathbf{f}_{i-1}\right] \in \{0,1\}^{(i-1)L}}$ consisting of the \acp{os} from all the preceding grasps. For the first grasp ($i=1$), we use a sentinel vector ${\mathbf{f}_\emptyset =
[-1,-1,-1,-1]}$ to signal the absence of history. $\mathbf{h}_{i-1}$ is appended to
the observation $\mathbf{o}_{t,i}$, enabling the policy to reason about already used \acp{dof}.

\subsection{Policy Architecture}
\methodname{} follows the encoder-denoiser architecture of \ac{dp3}~\citep{Ze2024DP3}. The noise-prediction network $\varepsilon_{\theta}$ is a 1D convolutional U-Net operating along the action-horizon dimension, following the standard \ac{dp} architecture~\citep{chi2024diffusionpolicy}. At every denoising step $k$, the U-Net input concatenates the \ac{os}-augmented noisy chunk $\tilde{\mathbf{A}}_t^{k}$, with the broadcast conditioning $\mathbf{c}_t$, and a sinusoidal embedding of $k$. 
\methodname{} is pre-trained using the behavior cloning objective in \equationref{eq:behavior_cloning_objective} on the dataset detailed in \secref{sec:dataset}. After pretraining, the same policy is \ac{rl} fine-tuned using \equationref{eq:dppo}, where the task-specific per-step reward $r_t^{\mathrm{base}}$ is introduced next.

\subsection{Reward for RL Fine-tuning}
\label{subsec:reward}

To make \methodname{} robust to states not covered by the \ac{il} dataset, we \ac{rl}-fine-tune it using \equationref{eq:dppo} on a dense, task-specific reward. The reward is shaped to encourage the policy to: (i) establish the selected \ac{os} contact on the current target object, (ii) keep every previously grasped object securely held, and (iii) avoid unnecessary off-\ac{os} contacts, idle-finger motion, and hand--table collisions that would compromise sim-to-real transfer. We encode these behaviors in the reward
\begin{equation}
r_t^{\mathrm{base}} =
\begin{cases}
r_{\mathrm{succ}}-\lambda_{\mathrm{ina}} r_t^{\mathrm{ina}},
& \text{if}~h_{i,t} = 1 \;\land\; c^{\mathrm{hold}}_{j,t} > 0 \;\;\forall j\in\mathcal{M}_i,\\[2pt]
\clip\!\left(
r_t^{\mathrm{grasp}}
+
r_t^{\mathrm{maintain}}
-
r_t^{\mathrm{avoid}},
r_{\min}, r_{\max}
\right),
& \text{otherwise}.
\end{cases}
\label{eq:reward_base}
\end{equation}

The policy is updated with a large constant reward $r_{\mathrm{succ}}$ at every step where the target object is lifted (normalized lift score $h_{i,t}=1$) and all previously grasped objects in $\mathcal{M}_i$ are securely held. Because this condition is re-evaluated at every step rather than terminating the episode, the policy is rewarded for both picking the new object and holding the earlier ones. The penalty $r_t^{\mathrm{ina}}$ discourages fingers that are neither active in $\mathbf{f}_i$ nor recorded in the grasp history $\mathbf{h}_{i-1}$ from drifting away from their initial pose, suppressing the idle-finger twitching that otherwise emerges under sparse reward signals and would interfere with subsequent grasps.

The reward when success has not yet been achieved is split into three terms: $r_t^{\mathrm{grasp}}$, $r_t^{\mathrm{maintain}}$, and $r_t^{\mathrm{avoid}}$, each encouraging different behaviors. The first of these terms, $r_t^{\mathrm{grasp}}$, rewards the active fingers in $\mathbf{f}_i$ for approaching and contacting the current target, providing a smooth gradient that pulls the hand toward the commanded \ac{os}. The second term, $r_t^{\mathrm{maintain}}$, rewards keeping all previously grasped objects securely held, in contact, and stationary. The reward activates only when the new grasp contacts the target object and is scaled down by the movement of the least well-supported previously grasped objects. The scaling prevents the policy from neglecting the previously grasped objects in favor of the current object. The last term, $r_t^{\mathrm{avoid}}$, penalizes collisions and near-collisions between the hand and the table, as well as contacts between the target object and fingers not in $\mathbf{f}_i$ before the gripper makes contact with the object. Clipping the sum to $[r_{\min}, r_{\max}]$ keeps any single term from dominating early in training. 

\section{Dataset Curation}
\label{sec:dataset}

We collect a trajectory-level sequential multi-object grasping dataset in Robosuite~\citep{zhu2020robosuite} using the logic summarized in \algoref{alg:stage_curation}. This algorithm sweeps every object $b\in\mathcal{B}$ paired with each of the four \acp{os} $\mathbf{f}_1,\ldots,\mathbf{f}_4$, repeating $N$ trials per pair. The stage-indexed initializer $\textsc{Init}_i$ returns an initial simulator state $\sigma_0$ and the prior grasp history $\mathbf{h}_{i-1}$. For Stage~1, $\textsc{Init}_1$ resets to an empty scene with a randomized pose of $b$ and an empty history, while, for Stage~$i{>}1$, $\textsc{Init}_i$ samples a successful trajectory from $\mathcal{D}_{i-1}$, resumes from its final state, recovers its history $\mathbf{h}_{i-1}$, and adds the new target $b$ to the scene. Pairs with $\mathbf{f}_k\in\mathbf{h}_{i-1}$ are skipped, since the same \ac{os} cannot be reused. The pose generator $\textsc{Pose}(b,\mathbf{f}_k)$ produces a heuristic grasp $\mathbf{g}$ from the object's dimensions and the hand's geometry, selecting a top-down approach for the finger--finger \acp{os} $\mathbf{f}_1$, $\mathbf{f}_2$, $\mathbf{f}_3$ and a side approach for the ring--palm \ac{os} $\mathbf{f}_4$. An \ac{osc} controller then executes the pre-grasp, closure, and lift, producing a trajectory $\tau$. A trial is deemed successful if the robot lifts the object $10\,\mathrm{cm}$ above the table and the hold predicate $\textsc{Hold}(\tau, \mathbf{h}_{i-1})$ confirms all previously grasped objects remain in hand.
\begin{wrapfigure}{r}{0.46\textwidth}
\begin{minipage}{0.46\textwidth}
\begin{algorithm}[H]
\caption{Stage-$i$ trajectory collection}
\label{alg:stage_curation}
\KwIn{$\mathcal{B}$, $N$, $\mathcal{D}_{i-1}$}
\KwOut{$\mathcal{D}_i$}
\ForEach{$b\in\mathcal{B},\ k\in\{1,\ldots,4\}$}{
    \For{$n \gets 1$ \KwTo $N$}{
        $(\sigma_0, \mathbf{h}_{i-1}) \gets \textsc{Init}_i(b,\, \mathcal{D}_{i-1})$\;
        \lIf{$\mathbf{f}_k \in \mathbf{h}_{i-1}$}{\Continue}
        $\mathbf{g} \gets \textsc{Pose}(b, \mathbf{f}_k)$\;
        $\tau \gets \textsc{OSC}(\sigma_0, \mathbf{g})$\;
        \lIf{$\Delta z_b(\tau) > 10\,\mathrm{cm} \,\land\, \textsc{Hold}(\tau, \mathbf{h}_{i-1})$}{
            $\mathcal{D}_i \gets \mathcal{D}_i \cup \{(\tau,\, \mathbf{h}_{i-1} \cup \{\mathbf{f}_k\})\}$
        }
    }
}
\end{algorithm}
\end{minipage}
\vspace{-5.5\baselineskip}
\end{wrapfigure}

We run \algoref{alg:stage_curation} with $\mathcal{B}$ containing 15 distinct objects across three primitive geometries (spheres, cylinders, and boxes), yielding 564 trajectories for Stage~1, 592 for Stage~2, and 525 for Stage~3. 

\section{Experiments}
\label{sec:experiments}
We experimentally evaluate \methodname{} in both simulation and the real world.
Our evaluation is designed to answer the following questions:
\begin{enumerate}
    \item \label{q:baselines} How well does \methodname{} perform compared to baselines?
    \item \label{q:ablation} What does each component of \methodname{} contribute to the overall performance?
    \item \label{q:realworld} Does \methodname{} transfer to real hardware?
\end{enumerate}

\subsection{Simulation Experiments}
\label{sec:sim_exp}

\paragraph{Baselines.} We compare \methodname{} against the following methods:
\textbf{BC-RNN}~\citep{mandlekar2022matters}, a recurrent behavior-cloning baseline that models temporal dependencies across observations;
\textbf{PPO}~\citep{schulman2017proximal}, a standard on-policy \ac{rl} method trained from scratch on the dense reward in \equationref{eq:reward_base}; \textbf{SeqDiffuser}~\citep{seqgrasp}, a grasp-pose generation method paired with a motion planner, which operates on full object point clouds at test time; \textbf{\methodname{}-BC}, a \ac{dp3}~\citep{Ze2024DP3} pre-trained on point cloud input without the DPPO fine-tuning;. For all \ac{il} and \ac{rl} baselines, observations are kept consistent with \methodname{}. SeqDiffuser receives the full object point cloud as per its original formulation.

\paragraph{Evaluation protocol.} 
The simulation experiments are conducted in Robosuite~\citep{zhu2020robosuite}. For each method, we evaluate all four \acp{os} across all grasp stages, running 20 trajectories per stage for a total of 240 trials per method. The object set consists of 15 objects spanning three primitive geometries: spheres, cylinders, and boxes. The target object's pose is randomized within $\pm 10$~cm in translation and $\pm 45^\circ$ in rotation. For stages 2 and 3, we initialize the robot with 1 or 2 grasped objects, respectively, from a randomly sampled successful trajectory in the dataset. This provides controlled and identical starting conditions for every method. To ensure a fair comparison, the random seed is kept consistent across all methods. A trial is deemed successful if the target object is lifted at least $10$~cm above the table while all previously grasped objects remain securely held.

\paragraph{Results.} \tabref{tab:sim_sr_pen} reports the per-stage and average success rates for all methods across all 15 evaluation objects. The results show that all methods degrade as the number of objects simultaneously held increases, reflecting fewer available grasp options. Across all methods, \methodname{} achieves the highest success rate at all evaluated stages, demonstrating that \ac{os}- and grasp-history-conditioned \acp{dp} effectively preserve and exploit kinematic redundancies across sequential grasps.

Additionally, the results indicate that BC-RNN and \methodname{}-BC achieve competitive Stage~1 performance but degrade substantially at Stages~2 and~3, where the absence of explicit \ac{os} conditioning results in unstructured finger use that compromises subsequent grasps. \methodname{} outperforms \ac{dp} at all stages, demonstrating the benefit of \ac{dppo} fine-tuning. \ac{ppo} struggles across all stages due to Allegro Hand's high-dimensional action space, making reward-driven exploration from scratch impractical without a pre-trained initialization. While SeqDiffuser generates geometrically valid grasp poses, its reliance on a decoupled motion planner introduces execution failures, particularly in Stages~2 and~3, where collision-free planning becomes increasingly difficult due to previously held objects. In conclusion, \methodname{} outperforms all baselines thanks to its condition values and training setup, thereby answering \hyperref[q:baselines]{question~\ref*{q:baselines}}.


\begin{table}[t]
    \centering
    \begin{adjustbox}{max width=0.98\linewidth}
        \begin{tabular}{lccccc}
            \toprule
            \textbf{Method} & \textbf{BC-RNN} & \textbf{PPO} & \textbf{SeqDiffuser}
            & \textbf{\methodname{}-BC} & \textbf{\methodname{}~(Ours)} \\
            \midrule
            Stage 1
            & $52.50 {\scriptstyle \pm 1.02}$
            & $0.00 {\scriptstyle \pm 0.00}$
            & $1.67$
            & $57.08 {\scriptstyle \pm 2.57}$
            & $\mathbf{75.00 {\scriptstyle \pm 3.68}}$ \\
            Stage 2
            & $11.67 {\scriptstyle \pm 2.12}$
            & $0.00 {\scriptstyle \pm 0.00}$
            & $0.00$
            & $\mathbf{50.00 {\scriptstyle \pm 3.06}}$
            & $49.58 {\scriptstyle \pm 1.18}$ \\
            Stage 3
            & $27.50 {\scriptstyle \pm 2.70}$
            & $0.00 {\scriptstyle \pm 0.00}$ 
            & $0.00$
            & $42.08 {\scriptstyle \pm 1.56}$
            & $\mathbf{45.00 {\scriptstyle \pm 2.70}}$ \\
            \midrule
            Average
            & $30.56 {\scriptstyle \pm 0.20}$
            & $0.00 {\scriptstyle \pm 0.00}$
            & $0.56$
            & $49.72 {\scriptstyle \pm 0.71}$
            & $\mathbf{56.53 {\scriptstyle \pm 1.37}}$ \\
            \bottomrule
        \end{tabular}
    \end{adjustbox}
    \vspace{4pt}
    \caption{Per-stage grasp success rates (\%) across four \ac{os} conditions in
    simulation. Results are reported as mean $\pm$ standard deviation over three seeds.
    Each seed evaluates 20 trials per \ac{os} condition, totaling
    80 episodes per stage and 240 episodes across all three stages for each method.
    Stage~$i$ requires the robot to grasp the $i$-th object while retaining all
    previously grasped objects.}
    \label{tab:sim_sr_pen}
\end{table}


\begin{table}[bp]
    \centering
    \begin{adjustbox}{max width=0.9\linewidth}
        \begin{tabular}{lccc}
            \toprule
            \textbf{Variant} & \textbf{Stage 1} & \textbf{Stage 2}
            & \textbf{Stage 3} \\
            \midrule
            \methodname{}-BC
            & $57.08 {\scriptstyle \pm 2.57}$
            & $\mathbf{50.00 {\scriptstyle \pm 3.06}}$
            & $\mathbf{42.08 {\scriptstyle \pm 1.56}}$ \\
            w/o Grasp History Context
            & $\mathbf{62.50 {\scriptstyle \pm 3.68}}$
            & $48.75 {\scriptstyle \pm 5.30}$
            & $33.75 {\scriptstyle \pm 2.70}$ \\
            \midrule
            \methodname{} (full)
            & $\mathbf{75.00 {\scriptstyle \pm 3.68}}$
            & $\mathbf{49.58 {\scriptstyle \pm 1.18}}$
            & $\mathbf{45.00 {\scriptstyle \pm 2.70}}$ \\
            \ac{dppo} w/o $r_t^{\mathrm{grasp}}$
            & $70.83 {\scriptstyle \pm 1.18}$
            & $47.50 {\scriptstyle \pm 2.04}$
            & $38.33 {\scriptstyle \pm 6.64}$ \\
            \ac{dppo} w/o $r_t^{\mathrm{maintain}}$
            & $70.00 {\scriptstyle \pm 3.68}$
            & $44.17 {\scriptstyle \pm 4.12}$
            & $42.50 {\scriptstyle \pm 1.77}$ \\
            \bottomrule
        \end{tabular}
    \end{adjustbox}
    \vspace{4pt}
    \caption{Ablation study success rates (\%) in simulation. Results are reported
    as mean ${\scriptstyle \pm}$ standard deviation over three seeds. Each variant removes one design component from \methodname{}.}
    \label{tab:ablation}
\end{table}

\paragraph{Ablation Study.} We also ablate \methodname{} to understand how the different design choices affect the performance. The results are reported in \tabref{tab:ablation}. Removing the \ac{os} condition causes the largest drop at Stages~2 and~3, confirming that explicit finger-allocation signals are essential for sequential grasping. Removing the grasp history context similarly degrades later stages, as the policy cannot adapt its motion to previously occupied fingers. The \ac{dppo} fine-tuning phase yields consistent gains across all stages, validating its role in bridging the gap between demonstration coverage and deployment robustness. Finally, the \ac{os}-conditioned reward proves critical during \ac{rl} fine-tuning: without it, the policy converges to morphologies that violate the intended opposition space. Together, these results highlight that all the proposed design choices improve the performance in different aspects, thereby answering \hyperref[q:ablation]{question~\ref*{q:ablation}}.

\subsection{Real-World Experiments}
\label{sec:real_exp}
\begin{wrapfigure}{r}{0.40\textwidth}
  \vspace{-1.0\baselineskip}
  \centering
  \begin{adjustbox}{max width=\linewidth}
    \begin{tabular}{lccc}
      \toprule
      \textbf{Success Rate} & \textbf{Stage 1} & \textbf{Stage 2} & \textbf{Stage 3} \\
      \midrule
      \methodname{}-BC      & 40.00          & 20.00          & 6.67 \\
      \methodname{}         & \textbf{57.78} & \textbf{26.67} & \textbf{20.00} \\
      \bottomrule
    \end{tabular}
  \end{adjustbox}
  \captionof{table}{Real Experiment Results.}
  \label{tab:real_exp}
  \vspace{0.5\baselineskip}
  \includegraphics[width=\linewidth]{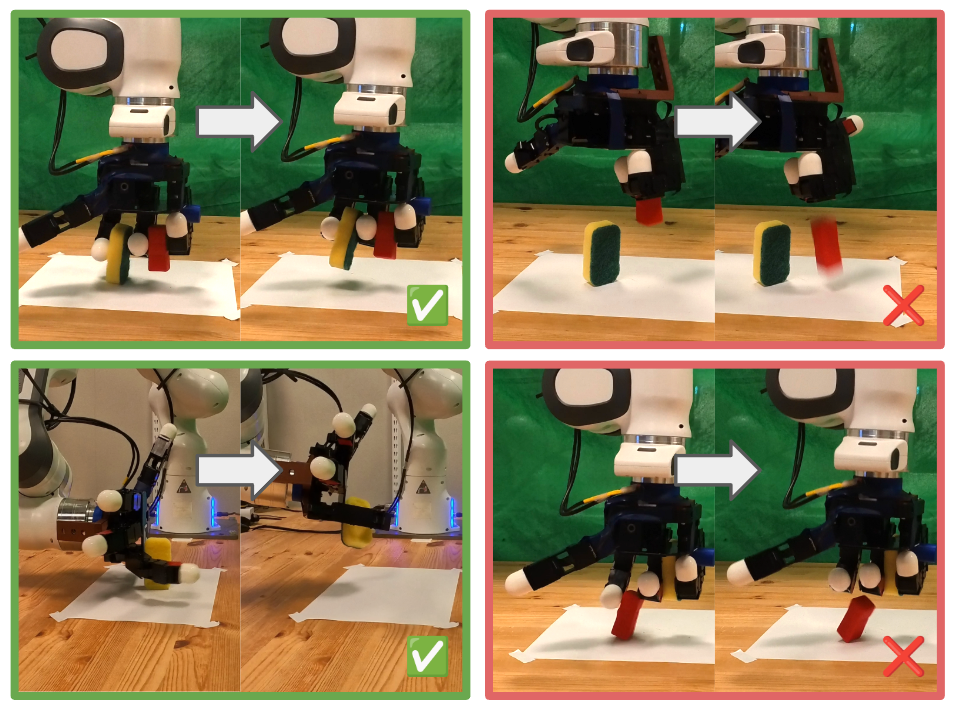}
  \captionof{figure}{Representative real-world rollouts. Two Successful sequences in green and the two failed ones in red. The failure case in the top image is due to the held object slipping out, while the bottom image is due to the Allegro Hand not closing tightly enough.}
  \label{fig:succ_fail_example}
\end{wrapfigure}
  
To assess real-world transferability, we deploy \methodname{} to control the real Allegro Hand and the \franka{} robot shown in \figref{fig:teaser}. We use a Kinect v3 depth sensor to capture the scene point cloud and follow the same evaluation protocol as in simulation, conducting 3 trials per object per stage. Our test set comprises three everyday objects and two 3D-printed objects, four of which are unseen during training. For stages~2 and~3, we prompt Gemini~\citep{geminiteam2024} to determine the object and $\mathbf{f}_i$ sequence. We use an \ac{osc} controller~\citep{osc} to move the arm to a predefined pose above the object before querying the policy and executing the resulting actions.

The real-world experimental results are reported in \tabref{tab:real_exp}. The results demonstrate that \methodname{} achieves promising zero-shot real-world transfer on out-of-distribution objects, with a modest drop relative to the simulation results, consistent with typical sim-to-real gaps in contact-rich manipulation~\citep{dexpoint,wei2025empirical,ren2025diffusion}. In contrast, \methodname{}-BC exhibits substantially greater sim-to-real degradation, particularly at Stages~2 and~3, strengthening the case that \ac{dppo} fine-tuning yields robust policies. A few different failure cases for \methodname{} are demonstrated in \figref{fig:succ_fail_example}, where the top image shows a previously grasped object slipping out of the hand, while the bottom image shows the finger coming very close to securing the target but fails to close around it due to the dynamic limits of the Allegro Hand. Despite these failure cases, the real-world results demonstrate that \methodname{}, although only trained on simulated data, can sequentially pick real-world objects, thereby answering \hyperref[q:realworld]{question~\ref*{q:realworld}}.

\section{Conclusion and Limitations}

This work explored the question: Are we underutilizing the \acp{dof} in dexterous grippers? To answer it, we introduced \methodname{}, a \ac{dp} trained to sequentially pick multiple objects one at a time by using a subset of the available \acp{dof}, while keeping all previously grasped objects securely held. The experimental evaluations demonstrated that the proposed \ac{os}- and grasp-history conditions, and the two-phase training scheme combining \ac{il} pre-training and \ac{rl} fine-tuning with a task-specific reward, were important factors behind \methodname{}'s performance. 

Despite the strengths, \methodname{} falls short of human-level dexterity in two key aspects. First, it does not support in-hand regrasping: humans routinely pinch an object with the thumb and index finger and then transfer it into a more secure power grasp. Picking up a pen is a perfect example. In contrast, \methodname{} commits to a single \ac{os} per object. Second, \methodname{} treats the \ac{os} assignment and the grasping order as given, rather than reasoning about which \ac{os} best suits each object or in what order objects should be picked. Addressing both would be a promising directionfor future work.


\bibliography{ref}  

\newpage

\end{document}